\documentclass{article} %
\usepackage{iclr2027_conference,times}

\usepackage{amsmath,amsfonts,bm}

\def\eqref#1{equation~\ref{#1}}

\def\1{\bm{1}}

\def\vk{{\bm{k}}}

\def\vp{{\bm{p}}}
\def\vq{{\bm{q}}}
\def\vr{{\bm{r}}}
\def\vs{{\bm{s}}}
\def\vt{{\bm{t}}}

\def\vv{{\bm{v}}}

\def\vx{{\bm{x}}}

\def\vz{{\bm{z}}}

\def\mA{{\bm{A}}}

\def\mD{{\bm{D}}}

\def\mI{{\bm{I}}}

\def\mM{{\bm{M}}}

\def\mP{{\bm{P}}}

\def\mR{{\bm{R}}}

\def\mV{{\bm{V}}}

\DeclareMathAlphabet{\mathsfit}{\encodingdefault}{\sfdefault}{m}{sl}
\SetMathAlphabet{\mathsfit}{bold}{\encodingdefault}{\sfdefault}{bx}{n}

\def\gD{{\mathcal{D}}}

\def\gJ{{\mathcal{J}}}

\def\gN{{\mathcal{N}}}
\def\gO{{\mathcal{O}}}
\def\gP{{\mathcal{P}}}

\def\gS{{\mathcal{S}}}

\def\sG{{\mathbb{G}}}

\newcommand{\E}{\mathbb{E}}
\newcommand{\Ls}{\mathcal{L}}

\newcommand{\KL}{D_{\mathrm{KL}}}

\newcommand{\TITLE}{{RAPP3R}\xspace}

\newcommand{\ontest}[1]{{\color{black} #1}}

\newcommand{\ie}{\textit{i.e.}\xspace}

\newcommand{\img}{\mI}
\newcommand{\depth}{\mD}
\newcommand{\pointmap}{\gP}
\newcommand{\dataset}{\gD}
\newcommand{\condition}{\mathcal{C}}

\newcommand{\pose}{\mP}
\newcommand{\mask}{\mM}
\newcommand{\shape}{\gS}
\newcommand{\scale}{\vs}
\newcommand{\rotation}{\mR}
\newcommand{\rotationsix}{\vr}
\newcommand{\translation}{\vt}

\newcommand{\object}{\gO}
\newcommand{\neighbor}{\gN}

\newcommand{\freq}{\omega}
\newcommand{\obj}{\gJ}

\newcommand{\pref}{p_\mathrm{ref}}
\newcommand{\ptheta}{p_\theta}
\newcommand{\pold}{p_{\theta_\mathrm{old}}}
\newcommand{\piref}{\pi_\mathrm{ref}}
\newcommand{\pitheta}{\pi_\theta}
\newcommand{\piold}{\pi_{\theta_\mathrm{old}}}

\newcommand{\sinc}{\mathrm{sinc}}
\newcommand{\clip}[1][\epsilon]{\mathrm{clip}_{1-#1}^{1+#1}}
\newcommand{\dist}{d}
\newcommand{\pen}{\delta}

\NewDocumentCommand{\nset}{O{i} O{N} m}{\{#3_{#1}\}_{#1=1}^{#2}}

\makeatletter
\newcommand{\trimcline}[1]{\@trimcline#1\@nil}
\def\@trimcline#1-#2\@nil{%
  \omit
  \@multicnt#1%
  \advance\@multispan\m@ne
  \ifnum\@multicnt=\@ne\@firstofone{&\omit}\fi
  \@multicnt#2%
  \advance\@multicnt-#1%
  \advance\@multispan\@ne
  \hskip\cmidrulekern
  \leaders\hrule\@height\cmidrulewidth\hfill
  \hskip\cmidrulekern
  \cr
  \noalign{\vskip-\cmidrulewidth}}
\makeatother

\definecolor{gating}{RGB}{250, 145, 141}
\definecolor{volume_rope}{RGB}{251, 175, 203}
\definecolor{reward}{RGB}{254, 152, 226}
\definecolor{ow_gdpo}{RGB}{189, 173, 214}

\newcommand{\first}{\cellcolor{red!30}}
\newcommand{\second}{\cellcolor{orange!30}}
\newcommand{\third}{\cellcolor{yellow!30}}

\usepackage{float}
\usepackage{hyperref}
\usepackage{url}
\usepackage{graphicx}
\usepackage{caption}
\usepackage{wrapfig}
\usepackage{cleveref}
\usepackage{booktabs}
\usepackage{tabularx}
\usepackage{bbm}
\usepackage{xspace}
\usepackage{verbatim}
\usepackage[table]{xcolor}

\title{RAPP3R: Reinforcing Object Relationship for Physically Plausible 3D Reconstruction}

\author{Seungwoo Yoon \& Jaeyun Jung \& Jaesik Park\thanks{Corresponding author.} \\
Seoul National University \\
\texttt{\{dotori000, cubecube1000, jaesik.park\}@snu.ac.kr}
}

\iclrfinalcopy %
\begin{document}

\maketitle
\lhead{Preprint}

\begin{abstract}
We present RAPP3R, a unified framework that combines architectural enhancements with a reinforcement learning strategy for physically plausible multi-object 3D scene reconstruction from a single image.
Building on a pre-trained joint shape-pose generator that processes each object independently, RAPP3R equips the model to reason about interactions across the objects.
Gated Inter-Object Attention introduces an inter-object branch while preserving the pre-trained model's behavior at initialization.
Volume-RoPE further encodes each object's position and spatial extent, providing explicit spatial context for attention across objects. 
Building on these architectural changes, we train the model with GDPO to learn physical relationships in a scene, such as collisions and floating.
We overcome the limitations of scene-level reinforcement learning through object-wise advantage computation.
As a result, our experiments show that RAPP3R outperforms prior methods in both pose accuracy and physical plausibility.
\end{abstract}

\section{Introduction}

Reconstructing real-world scenes requires capturing both object geometry and physically consistent spatial relationships to support applications such as robotic manipulation, physical simulation, and interactive digital twins.
Recovering such representations from one or a few images is challenging because occlusion and depth ambiguity leave geometry and placement underdetermined, while the reconstructed scene must remain visually faithful and physically plausible.
Recent generative approaches~\citep{SAM3D, RecGen} recover object shapes and spatial configurations, but object-level plausibility does not ensure scene-level physical consistency.
Small placement errors can break contact, leave objects floating, or introduce inter-object penetration, limiting the usefulness of the reconstructed environment.

Physical consistency depends on how objects are arranged relative to one another, motivating explicit modeling of inter-object relationships.
Inter-object attention~\citep{MIDI, SceneGen} enables information exchange across objects, while physics-aware methods explicitly correct generated scenes through contact-guided rejection sampling~\citep{PICASSO} or joint shape-pose optimization~\citep{SPARCS}.
However, these correction procedures require additional scene-specific computation at inference time.
To reduce reliance on such correction and incorporate physical consistency into the generator, we turn to reinforcement learning~(RL) with geometric and physical rewards.

To this end, we introduce \textbf{\TITLE}, \textbf{R}einforcing Object Rel\textbf{A}tionship for \textbf{P}hysically \textbf{P}lausible \textbf{3}D \textbf{R}econstruction, a unified architecture and training framework for multi-object 3D reconstruction from a single image.
\TITLE couples explicit modeling of inter-object relationships with object-wise reward assignment to learn physically consistent object arrangements.
\textit{Gated Inter-Object Attention} enables information exchange across objects while preserving the pretrained model's behavior at initialization.
\ontest{\textit{Volume-RoPE} encodes both object positions and spatial extents, providing spatial context for inter-object attention.}
We train \TITLE through reinforcement learning, using geometric and physical rewards inspired by physics-aware optimization~\citep{PICASSO, SPARCS}.
To distinguish the quality of individual objects within a generated scene, we introduce \textit{Object-wise GDPO}, which extends GDPO's reward-wise normalization~\citep{GDPO} to individual objects and assigns each object's pose update a separate advantage.

Our experiments demonstrate that \TITLE reconstructs 3D scenes with improved physical consistency and better alignment with the ground truth compared with previous methods. Our main contributions are:

\begin{itemize}
\item We adapt a pre-trained joint shape-pose generator to model inter-object relationships through \textit{Gated Inter-Object Attention}, preserving its original behavior at initialization.
\item We derive \textit{Volume-RoPE} by averaging rotary positional embeddings over object bounding boxes, encoding both object positions and spatial extents for inter-object attention.
\item We develop \textit{Object-wise GDPO} for reinforcement learning with geometric and physical rewards, extending GDPO's reward-wise normalization~\citep{GDPO} to individual objects to provide object-specific reward assignment.
\end{itemize}

\section{Related Work}

\subsection{Generative 3D Reconstruction}

Generative 3D reconstruction uses learned generative priors to recover 3D geometry and appearance from visual observations.
Many image-to-3D generators~\citep{VecSet, TripoSG, Hunyuan3D, TRELLIS, TRELLIS2, Direct3DS2} employ diffusion~\citep{DDPM} or flow matching~\citep{Flow} in latent spaces and primarily focus on individual objects.
Composing these objects into a scene additionally requires predicting their scales, rotations and translations.
MegaPose~\citep{MegaPose}, SAM-6D~\citep{SAM6D}, and FoundationPose~\citep{FoundationPose} estimate these 6D poses using supplied object models or additional reference observations.

To recover geometry and placement within a unified framework, recent generative approaches jointly infer object shapes and spatial configurations from scene observations.
SAM3D~\citep{SAM3D}, RecGen~\citep{RecGen}, and CUPID~\citep{CUPID} jointly model canonical object geometry and the relative pose between the object and camera.
MIDI~\citep{MIDI} and SceneGen~\citep{SceneGen} further model inter-object relationships through attention mechanisms to generate spatially coherent scenes.
Despite these advances in geometric reconstruction and spatial coherence, physical artifacts such as inter-object penetration and floating objects persist.
We address these challenges by aligning multi-object scene generation with geometric and physical objectives through explicit object-wise reward assignment.

\subsection{Physics-aware 3D Scene Reconstruction}

Physics-aware methods often augment pretrained reconstruction models such as SAM3D~\citep{SAM3D} with additional inference-time sampling or optimization to explicitly enforce physical constraints.
PICASSO~\citep{PICASSO} uses an inferred contact graph to guide rejection sampling under contact and non-penetration constraints.
SPARCS~\citep{SPARCS} and CA-World~\citep{CAWorld} incorporate physical constraints or penalties into scene-specific optimization.
MV-SAM3D~\citep{MV-SAM3D} applies collision and contact guidance during structure generation in a multi-view setting, followed by global pose refinement.

Other approaches use simulation feedback to guide scene correction.
REST3D~\citep{REST3D} evaluates candidate layouts through simulation within hierarchical optimization, while $\phi$-Scene~\citep{PhiScene} optimizes object positions to resolve penetrations and uses rigid-body simulation to settle objects in an order determined by their support relationships.
SimuScene~\citep{SimuScene} uses diagnostic simulation to guide gravity-axis shape stretching and amodal shape resampling.
Complementing these inference-time approaches, our method uses object-wise geometric and physical rewards to improve the flow-matching generator during training.

\subsection{Alignment of 3D Generative Models}

Reinforcement learning enables generative models to optimize externally evaluated objectives without differentiating through the evaluator, as demonstrated by large language model alignment with human preferences~\citep{RLHF} and verifiable rewards~\citep{DeepSeekMath}.
Within this paradigm, Group Relative Policy Optimization (GRPO)~\citep{DeepSeekMath} estimates advantages from relative rewards among sampled outputs, while Group reward-Decoupled Normalization Policy Optimization (GDPO)~\citep{GDPO} separately normalizes reward components to accommodate multiple objectives.
These approaches have also been extended to denoising models, with Diffusion-DPO~\citep{DiffusionDPO} adapting Direct Preference Optimization (DPO)~\citep{DPO} to diffusion models and Flow-GRPO~\citep{FlowGRPO} enabling online reward optimization for flow matching models through SDE sampling.

For 3D generation, DSO~\citep{DSO} adapts Diffusion-DPO~\citep{DiffusionDPO} to fine-tune single-object generators using simulator-derived stability feedback.
SimuScene~\citep{SimuScene} uses flow-matching DPO to improve occlusion-robust amodal shape completion within a simulation-guided reconstruction pipeline.
For multi-object generation, PhyMix~\citep{PhyMix} introduces Scene-GRPO, which reweights scene-level flow-matching losses using physics-evaluator rewards and combines this training with test-time refinement.
However, Scene-GRPO assigns a shared advantage to all objects within each candidate scene, providing no separate reward for well-placed and poorly placed objects.
Our work extends GDPO's reward-wise normalization to individual objects, assigning each object's pose update a separate advantage derived from geometric and physical rewards within a jointly generated scene.

\section{Method}

\begin{figure*}
    \centering
    \includegraphics[width=\textwidth]{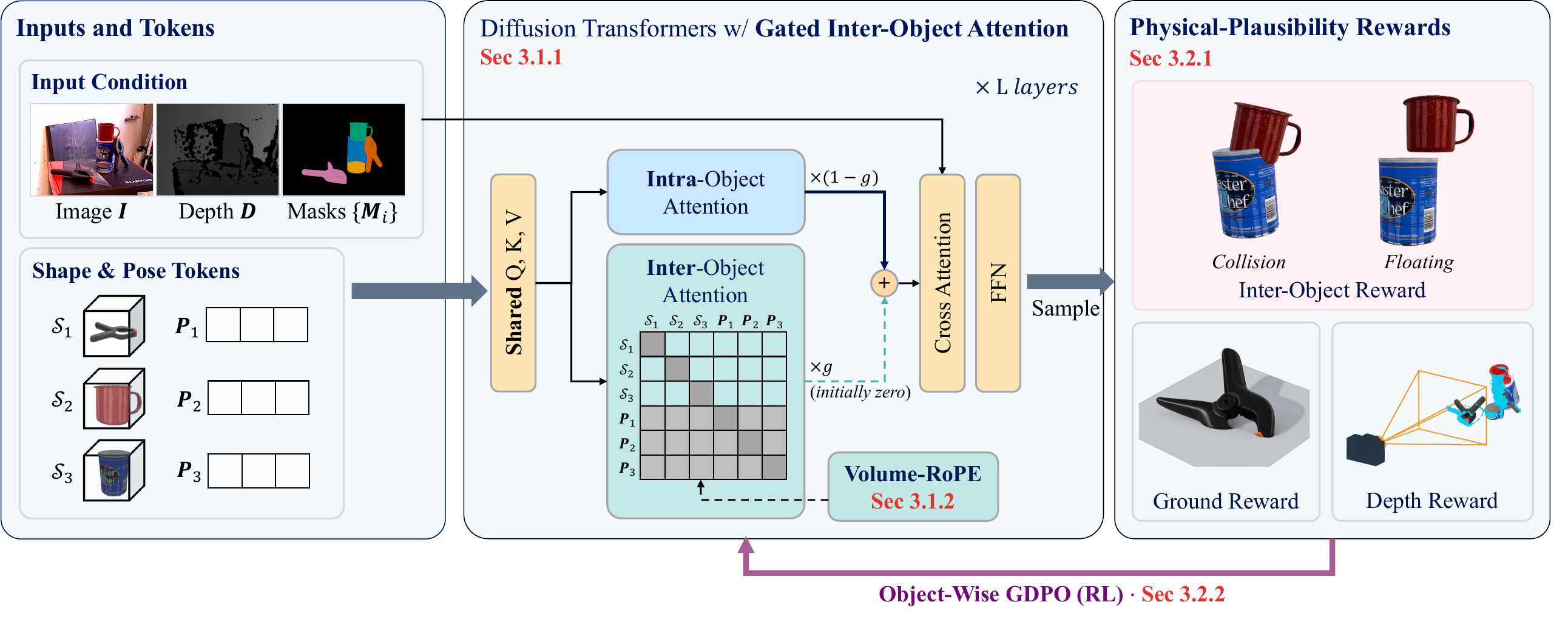}
    \captionof{figure}{\textbf{The overall pipeline of \TITLE.}
    \TITLE integrates gated inter-object attention into a pretrained model to capture object relationships~(\cref{sec:io_attn}), with Volume-RoPE encoding object positions and spatial extents~(\cref{sec:volume_rope}).
    We train the model using geometric and physical rewards~(\cref{sec:rewards}) with Object-wise GDPO for object-specific reward assignment~(\cref{sec:ow_gdpo}).
    }
    \vspace{-5mm}
    \label{fig:io_attn}
\end{figure*}

\TITLE jointly generates multiple 3D objects from a single image, recovering their geometries and spatial arrangement.
Given an input image $\img$, a depth map $\depth$, and object masks $\{\mask_i\}_{i=1}^{N}$, the model predicts each object's shape $\shape_i$ and pose $\pose_i=(\scale_i,\rotation_i,\translation_i)$, where $\scale_i$, $\rotation_i$, and $\translation_i$ denote scale, rotation, and translation, respectively.

Our method builds on SAM3D~\citep{SAM3D}, a pre-trained flow-based joint shape-pose estimator $\pref$ that generates each object's shape and pose independently: \ie $\pref\!\left(\nset{(\shape,\pose)} \mid \img,\depth,\{\mask_i\}_{i=1}^{N}\right)=\prod_{i=1}^{N}\pref\!\left(\shape_i,\pose_i \mid \img,\depth,\mask_i\right)$.
We present a unified framework for model architecture~(\cref{sec:architecture}) and training~(\cref{sec:training}) to improve physical plausibility, with the overall pipeline illustrated in \cref{fig:io_attn}.

\subsection{Architecture}
\label{sec:architecture}

\subsubsection{Gated Inter-Object Attention}
\label{sec:io_attn}

To achieve scene-level physical plausibility, it is necessary to consider physical relationships between objects, such as penetration and floating.
We therefore introduce inter-object attention inspired by prior work on multi-object generation~\citep{MIDI, SceneGen}.
Following MIDI~\citep{MIDI}, we reuse pre-trained model parameters for inter-object attention.
The object pose tokens in the original SAM3D can attend to the shape and pose tokens of the same object, which we refer to as intra-object attention.
In contrast, inter-object attention allows pose tokens to attend to the shape and pose tokens of other objects.

To incorporate inter-object attention into the model, MIDI selectively replaces some intra-object attention layers with their corresponding inter-object attention layers.
However, this replacement leads to an initial performance degradation because the layers were originally trained on intra-object attention rather than inter-object attention.
Inspired by prior work~\citep{LLaMA-Adapter, Video-LDM} on attaching new modules to pre-trained models to impart new functionality, we introduce a gating mechanism~(named after \citealt{LLaMA-Adapter}) that interpolates between intra-object and inter-object attention.
Instead of the original attention, a weighted average of intra-object and inter-object attention is used with shared parameters and a trainable weight $g\in[-1, 1]$.
By initializing $g$ to 0, all attention layers initially operate as intra-object attention and are gradually trained to consider inter-object relationships.

\ontest {
\subsubsection{Volume-RoPE}
\label{sec:volume_rope}

When shape and pose tokens from multiple objects interact, spatial cues such as adjacency can help the model relate them to one another.
Across diverse transformer architectures, positional encodings provide such spatial context.
Rotary Positional Encoding~(RoPE)~\citep{Roformer} has recently become a widely adopted choice for this purpose.

Conceptually, for a token at position $x$, RoPE rotates its query and key representations by $\freq x$ radians (\ie multiplies them by $e^{i\freq x}$), where $\freq$ is the frequency associated with the corresponding vector dimension.
A naive extension of RoPE to multi-object 3D generation would use each object's center as its RoPE position; we call this formulation as Center-RoPE (\cref{fig:volume_rope}).
However, center locations alone cannot fully characterize spatial relationships
between objects, as relationships such as adjacency also depend on
their spatial extents.

We therefore derive Volume-RoPE, which represents both the position and volume of an object. 
Consider arbitrary points $x$ in object $i$ and $y$ in object $j$.
Under standard RoPE, their relative position is embedded as $e^{i\freq(x-y)}$.
We approximate the mean relative embedding between the two objects by averaging over their bounding boxes:
\begin{equation}
\label{eq:volume_rope}
\E[e^{i\freq(x-y)}]=\frac{1}{4h_ih_j}\int_{t_j-h_j}^{t_j+h_j}\int_{t_i-h_i}^{t_i+h_i}e^{i\freq(x-y)}\mathrm{d}x\mathrm{d}y=\sinc(\freq h_i)\sinc(\freq h_j)e^{i\freq(t_i-t_j)},
\end{equation}
where $t_i$ and $h_i$ denote the center position and half-extent of object $i$, respectively.
Thus, the positional encoding of object $i$ can be represented as $\sinc(\freq h_i)e^{i\freq t_i}$, a computationally feasible form, which we call as Volume-RoPE.
The derivation is illustrated intuitively in \cref{fig:volume_rope}; we defer the mathematical details to \cref{sec:rope_math}.

\begin{wrapfigure}{l}{0.45\textwidth}
    \vspace{-2.5mm}
    \centering
    \includegraphics[width=0.45\textwidth]{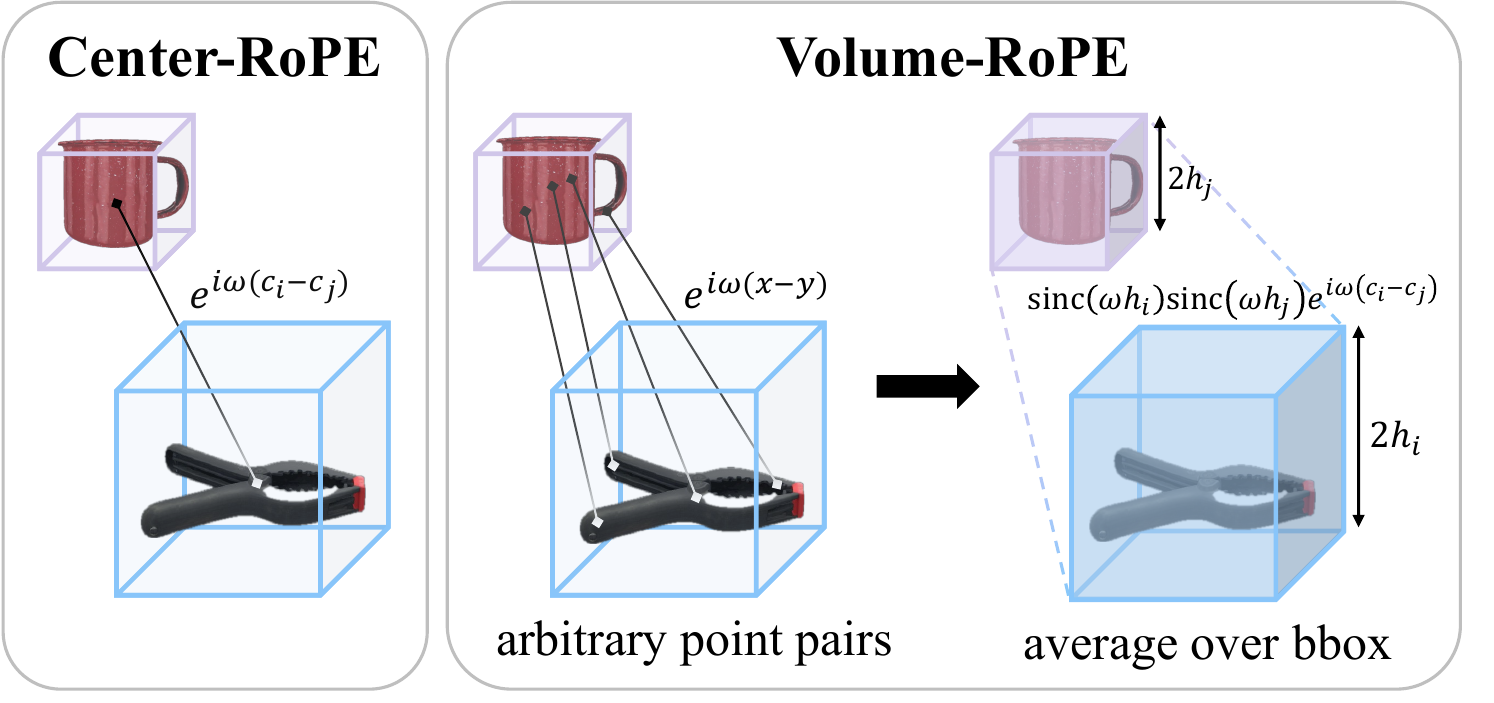}
    \vspace{-5mm}
    \captionof{figure}{\textbf{Volume-RoPE.}
    Unlike Center-RoPE, Volume-RoPE encodes the spatial extents of generated objects.
    }
    \label{fig:volume_rope}
\end{wrapfigure}

A challenge is that Volume-RoPE requires object positions and spatial extents, which are unknown before generation.
We approximate these quantities from a pointmap $\pointmap$ derived from the depth map.
For each object, we construct a cubic bounding box from the masked pointmap $\mask_i \odot \pointmap$ and denote its center and half-extent by $c_i$ and $h_i$, respectively.
To achieve invariance to scene-level scaling and translation, we subtract the mean of $\nset{c}$ from the centers and divide both the centered positions and the half-extents by the standard deviation of $\nset{c}$.
Using these values, we replace $t_i$ with $c_i$ to obtain the Volume-RoPE encoding $\sinc(\freq h_i)e^{i\freq c_i}$.

}

\subsection{Training}
\label{sec:training}

We introduce four types of reward to inject physical plausibility into the pre-trained model: (1)~inter-object reward~($\mathrm{io}$), (2)~ground reward~($\mathrm{grd}$), (3)~depth reward~($\mathrm{depth}$), and (4)~GT reward~($\mathrm{gt}$). 
Throughout this section, the scalar $r_{i,a,b}$ denotes the reward of type $b\in \{\mathrm{io},\mathrm{grd},\mathrm{depth},\mathrm{gt}\}$ for object $a$
in sample $i$.

\subsubsection{Physical-Plausibility Rewards}
\label{sec:rewards}

\noindent{\textbf{Inter-object reward.}}
Let $\object_{i,a}$ denote the posed object $a$ in sample $i$.
The inter-object reward is
\begin{equation}
\label{eq:io_reward}
r_{i,a,\mathrm{io}}
= -\frac{1}{N-1}
    \sum_{\substack{k=1\\k\ne a}}^{N}
    \pen(\object_{i,a},\object_{i,k})
  -\frac{1}{|\neighbor_a|}
    \sum_{k\in\neighbor_a}
    \dist(\object_{i,a},\object_{i,k}).
\end{equation}
\begin{wrapfigure}{r}{0.25\textwidth}
    \vspace{-5mm}
    \centering
    \includegraphics[width=0.25\textwidth]{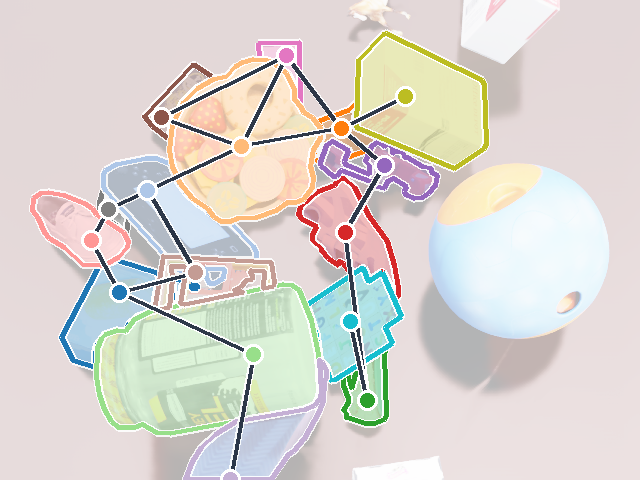}
    \vspace{-5mm}
    \captionof{figure}{Contact Graph.}
    \label{fig:contact_graph}
\end{wrapfigure} 
Here, $\pen(A,B)\geq 0$ and $\dist(A,B)\geq 0$ denote the penetration depth and distance between posed objects $A$ and $B$, respectively. $\neighbor_a$ contains the indices of objects in contact with object $a$ in the ground-truth contact graph (\cref{fig:contact_graph}). The two terms penalize inter-object penetration and separation between objects that should be in contact, respectively.

\noindent{\textbf{Ground reward.}}
Let $z(\object_{i,a}) := \min_{\vp\in\object_{i,a}} \vp_z$
denote the lowest vertical coordinate of object $a$ in sample $i$.
Assume that the ground plane is at $z=0$, the ground reward is
\begin{equation}
\label{eq:grd_reward}
r_{i,a,\mathrm{grd}}
= \min(0,z(\object_{i,a}))
- \mathbbm{1}_{a\in\sG}\max(0,z(\object_{i,a})),
\end{equation}
where $\sG$ indexes objects touching the ground in the ground-truth scene.
The first term penalizes ground penetration for every object,
while the second penalizes floating for objects in $\sG$.

\noindent{\textbf{Depth reward.}}
This reward discourages predicted objects from occupying observed empty space.
Inspired by the free-space constraint in PICASSO~\citep{PICASSO},
the depth reward penalizes object regions that are rendered
in front of the observed depth map:
\begin{equation}
\label{eq:depth_reward}
r_{i,a,\mathrm{depth}}
= -\sum_{\vx \in \mathrm{render}(\object_{i,a})}
  \max(0,\depth_\vx-\hat\depth_{i,\vx}),
\end{equation}
where $\vx$ indexes an image pixel,
$\depth_\vx$ is the observed depth at that pixel, 
$\hat\depth_{i,\vx}$ is the depth rendered from the predicted scene in sample $i$
, and $\mathrm{render}(\object_{i,a})$ denotes the set of pixels
with valid observed depth at which object $a$ is visible
in the rendered scene.
When $\hat\depth_{i,\vx}<\depth_\vx$, \ie when part of an object
lies closer to the camera than the observed surface,
the object receives a negative reward.

\noindent{\textbf{GT reward.}} 
The GT reward regularizes the predicted object poses toward
their ground-truth values during RL fine-tuning.
It is defined as the negative $\ell_2$ distance between the
predicted and ground-truth pose parameters, comprising
scale $\scale_a$, 6D rotation
representation $\rotationsix_a$, and translation $\translation_a$:
\begin{equation}
\label{eq:gt_reward}
r_{i,a,\mathrm{gt}}
= -\left\|
(\log\hat\scale_{i,a},
 \hat\rotationsix_{i,a},
 \hat\translation_{i,a})
-
(\log\scale_a,\rotationsix_a,\translation_a)
\right\|_2,
\end{equation}

\subsubsection{Object-Wise GDPO}
\label{sec:ow_gdpo}

The equations below describe an application of Flow-GRPO~\citep{FlowGRPO} to a flow-based pose estimator with conditioning inputs $\condition=(\img,\depth,\{\mask_a\}_{a=1}^{N},\{\shape_a\}_{a=1}^{N})$, comprising the input image, depth map, object masks, and ground-truth object shapes:
\begin{equation}
\label{eq:objective}
\obj(\theta)=\E_{\condition\sim\dataset, \nset[i][G]{\vx}\sim\piold(\cdot\mid\condition)}[f(\theta)],
\end{equation}
\begin{equation}
\begin{aligned}
f(\theta)=\frac{1}{GT}\sum_{i=1}^{G}\sum_{t=1}^{T}
\Big[
\min\left(
s_i^t(\theta)\hat A_i,\,
\clip(s_i^t(\theta))\hat A_i
\right)
-\beta D_{\mathrm{KL}}\left(
\pitheta\,\middle\|\,\piref
\right)
\Big].
\end{aligned}
\end{equation}
\begin{equation}
s_i^t(\theta)=
\frac{\ptheta(\vx_i^{t-1}\mid\vx_i^t,\condition)}
     {\pold(\vx_i^{t-1}\mid\vx_i^t,\condition)}.
\end{equation}
Here, $\dataset$ is the training-data distribution, and $\pitheta$ denotes the stochastic denoising policy parameterized by $\theta$, with one-step transition density $\ptheta$.
Each $\vx_i=(\vx_i^T,\ldots,\vx_i^0)$ is a scene-level pose-estimation trajectory obtained by rolling out the behavior policy $\piold$ for $T$ denoising steps.
The objective $\obj(\theta)$ is maximized using the group-relative advantage $\hat A_i$ computed from $G$ samples.
Details of the stochastic policy and KL computation are provided in \cref{sec:rl}.

We fine-tune only the pose-estimation component of the pre-trained model while fixing the shape-generation latent
at step $t$ to $\vz^t=\frac{T-t}{T}\vz^0+\frac{t}{T}\boldsymbol{\xi}$, where $\vz^0$ is the ground-truth shape latent and
$\boldsymbol{\xi}\sim\mathcal{N}(\mathbf{0},\mathbf{I})$ is Gaussian noise.

To prevent reward components with larger scales or variances from dominating the learning signal, GDPO~\citep{GDPO} employs two-step normalization to get the advantage:
\begin{equation}
A_{i, b} = \frac{r_{i, b} - \mathrm{mean}\{r_{j, b}\}_{j=1}^G}{\mathrm{std}\{r_{j, b}\}_{j=1}^G}, \quad
A_i = \sum_{b=1}^M A_{i, b}, \quad
\hat A_i = \frac{A_i - \mathrm{mean}\nset[j][G]{A}}{\mathrm{std}\nset[j][G]{A}},
\end{equation}
where $b$ indexes the reward type and $M$ is the number of reward types.
GDPO first normalizes rewards within each reward type, sums the normalized rewards across types, and then normalizes the aggregate scores across samples to obtain $\hat A_i$.

\begin{wrapfigure}{l}{0.61\textwidth}
    \centering
    \vspace{-3mm}
    \includegraphics[width=\linewidth]{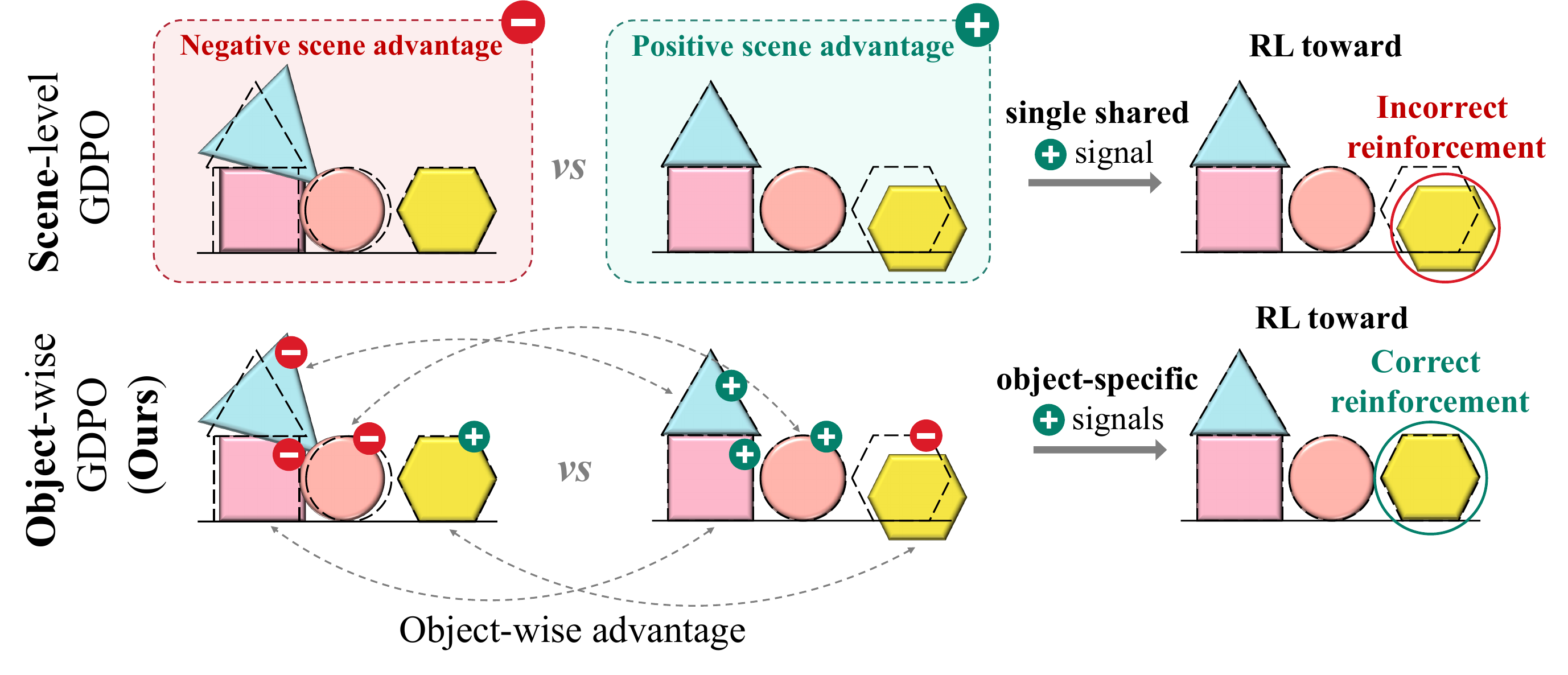}
    \vspace{-7.5mm}
    \caption{Comparison of Scene-level GDPO and Object-wise GDPO. }
    \label{fig:object_wise_gdpo}
\end{wrapfigure}
However, directly applying GRPO or GDPO to multi-object 3D generation cannot distinguish quality differences among individual objects because all objects share a scene-level advantage, as in the Scene-GRPO formulation of PhyMix~\citep{PhyMix}.
\Cref{fig:object_wise_gdpo} illustrates how an incorrectly placed object in a scene with a positive advantage can be reinforced, whereas a correctly placed object in a scene with a negative advantage can be penalized.

To address this issue, we extend GDPO's reward-decoupling strategy to the object level:
\begin{equation}
A_{i, a, b} = \frac{r_{i, a, b} - \mathrm{mean}\{r_{j, a, b}\}_{j=1}^G}{\mathrm{std}\{r_{j, a, b}\}_{j=1}^G}, \quad
A_{i, a} = \sum_{b=1}^M A_{i, a, b},
\end{equation}
where $a$ indexes an object in a scene. The decoupled advantages are then normalized along the group axis:
\begin{equation}
\hat A_{i, a} = \frac{A_{i, a} - \mathrm{mean}\{A_{j, a}\}_{j=1}^{G}}{\mathrm{std}\{A_{j, a}\}_{j=1}^{G}}.
\end{equation}
Finally, we compute the object-wise objective $g(\theta)$ using the object-decoupled advantage $\hat A_{i, a}$:
\begin{equation}
g(\theta)=\frac{1}{GT}\sum_{i=1}^{G}\sum_{t=1}^{T}
\Biggl\{
\sum_{a=1}^{N}\Bigl[
\min\bigl(
s_{i,a}^t(\theta)\hat A_{i,a},
\clip(s_{i,a}^t(\theta))\hat A_{i,a}
\bigr)
\Bigr]
-\beta\KL(\pitheta\|\piref)
\Biggr\}.
\end{equation}
\begin{equation}
s_{i,a}^t(\theta)=\frac{\ptheta(\vx_{i,a}^{t-1}\mid\vx_i^t,\condition)}{\pold(\vx_{i,a}^{t-1}\mid\vx_i^t,\condition)}.
\end{equation}

Replacing $f(\theta)$ with $g(\theta)$ in \cref{eq:objective}
yields our object-wise RL objective.
The final training loss combines this KL-regularized objective
with the SFT loss defined in \cref{sec:sft}:
\begin{equation}
\label{eq:loss}
\Ls(\theta)=-\obj(\theta)+\lambda\Ls_{\mathrm{SFT}}(\theta).
\end{equation}

\section{Experiments}

\begin{figure*}
    \centering
    \includegraphics[width=\textwidth]{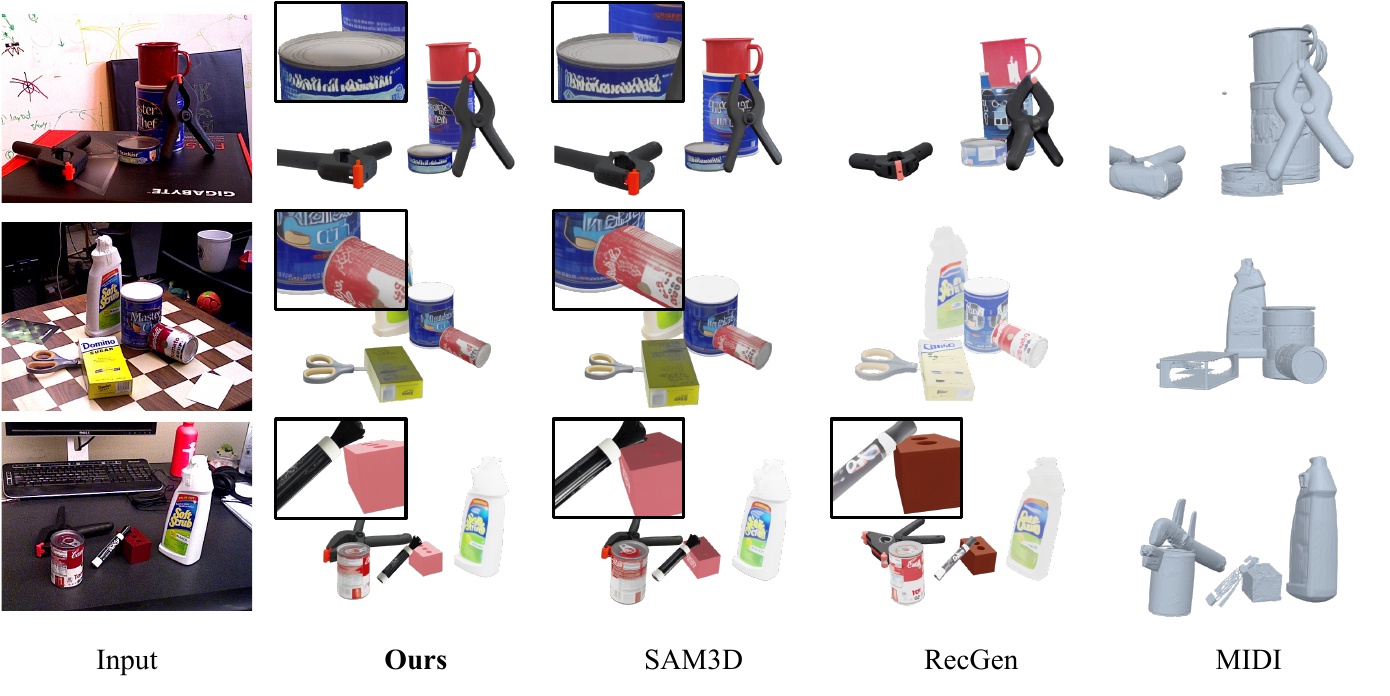}
    \captionof{figure}{\textbf{Qualitative Results.}
    The output 3D scenes are rendered in the original camera view of input images.
    The parts where our method does not violate physics while others show collision are emphasized.
    }
    \label{fig:results}
\end{figure*}

\subsection{Setup}
\label{sec:setup}

\textbf{Datasets.}
We train on the dataset introduced by FoundationPose~\citep{FoundationPose}, which comprises synthetic scenes with multiple objects from GSO~\citep{GSO}.
This dataset is well suited for learning inter-object interactions because it captures diverse physical relationships, including support, contact, and occlusion.
We evaluate the generated poses on the YCB-V~\citep{PoseCNN}, a widely used benchmark for 6D object pose estimation.
The real-world scenes in YCB-V likewise span a broad range of physical configurations and interactions.
The benchmark comprises 12 such scenes with 75 frames each, yielding a total of 900 inputs.

\textbf{Training.}
For GDPO training, we use size of a group $G=16$.
Coefficients of regularization terms in \cref{eq:loss} are set to be $\beta=1.0$ and $\lambda=1.0$ empirically.
During group sampling, we set inference step $T=10$ following the recipe of Flow-GRPO.
While learning rate is basically $\gamma=4\times10^{-4}$, the learning rate for gating coefficient $g$ is 50 times larger so that the model can utilize inter-object attention more quickly.
We finetune the model using LoRA~\citep{LoRA} with rank 32.
We trained the model using 500 scene samples of the training data.

\textbf{Metrics.}
ADD-S~\citep{PoseCNN} is the most commonly used metric for evaluating pose estimation accuracy.
We use the SAM3D variant of ADD-S, which accounts for both shape-generation and pose-estimation accuracy.
We also report ADD-S@0.1, defined as the fraction of object predictions with normalized ADD-S below 0.1.
To evaluate physical plausibility, we report the average collision ratio~(Collision) as a percentage~(\%) and the average floating distance~(Floating) and ground displacement~(Ground) in millimeters~(mm).
The inter-object collision ratio is the fraction of object pairs in which one object penetrates more than 5\% of the other.
Floating distance is zero for colliding objects and otherwise measured to the nearest object that has a lower center and shares support.
For objects touching the ground in the ground-truth scene, ground displacement measures the distance from their bottom to the ground plane.

\textbf{Baselines.}
We compare our method against three training-based joint shape-and-pose generation methods: MIDI~\citep{MIDI}, RecGen~\citep{RecGen}, and SAM3D~\citep{SAM3D}.
We additionally evaluate SAM3D combined with the optimization-based method PICASSO~\citep{PICASSO}.
For a direct comparison under the same refinement protocol, we also apply PICASSO to our method.

\subsection{Main Results}

\begin{table}
\caption{\textbf{Results.} Quantitative comparison of our approach and baselines.}
\vspace{-5mm}
\label{tab:results}
\begin{center}
\footnotesize
\setlength{\tabcolsep}{1pt}
\begin{tabularx}{\linewidth}{@{}l<{\hspace{4pt}}*{5}{>{\centering\arraybackslash}X}@{}}
    \toprule
     & \multicolumn{2}{c}{Alignment} & \multicolumn{3}{c}{Physics} \\
    \trimcline{2-3}\trimcline{4-6}
     Method 
    & ADD-S $\downarrow$ & \mbox{ADD-S@0.1$\!\uparrow$}
    & \rule{0pt}{2.4ex} Collision $\downarrow$ & Floating $\downarrow$
    & Ground $\downarrow$ \\
\midrule
MIDI~\citep{MIDI}
& 0.3372 & 0.0989 & 17.23 & 5.36 & \first 4.27 \\
RecGen~\citep{RecGen}
& \third 0.0780 & \second 0.9329 & \second 8.37 & \third 2.24 & \third 9.97 \\
\midrule
SAM3D~\citep{SAM3D}
& \second 0.0697 & \third 0.8962 & \third 10.24 & 3.35 & \second 9.47 \\
SAM3D (\textit{repo ver.})
& 0.0910 & 0.7724 & 18.13 & \second 1.63 & 16.66 \\
\midrule
\textbf{\TITLE (Ours)}
& \first 0.0685 & \first 0.9556 & \first 5.01 & \first 1.50 & 11.20 \\
\bottomrule
\end{tabularx}
\rightline{* Explanation about \textit{repo ver.} is in \cref{sec:use_scene_scale}.}
\end{center}
\end{table}

\begin{table}
\caption{\textbf{Post-Processing Results.}}
\vspace{-5mm}
\label{tab:post_results}
\begin{center}
\footnotesize
\setlength{\tabcolsep}{1pt}
\begin{tabularx}{\linewidth}{@{}l<{\hspace{4pt}}*{5}{>{\centering\arraybackslash}X}@{}}
    \toprule
     & \multicolumn{2}{c}{Alignment} & \multicolumn{3}{c}{Physics} \\
    \trimcline{2-3}\trimcline{4-6}
     Method 
    & ADD-S $\downarrow$ & \mbox{ADD-S@0.1$\!\uparrow$}
    & \rule{0pt}{2.4ex} Collision $\downarrow$ & Floating $\downarrow$
    & Ground $\downarrow$ \\
    \midrule
    SAM3D + PICASSO~\citep{PICASSO} & 0.0492 & 0.9707 & 11.09 & \textbf{2.90} & 7.82 \\
    \textbf{\TITLE (Ours)} + PICASSO & \textbf{0.0482} & \textbf{0.9782} & \textbf{9.34} & 4.21 & \textbf{5.84} \\
    \bottomrule
\end{tabularx}
\end{center}
\end{table}

\Cref{tab:results} reports the shape-pose accuracy and physical plausibility of the scenes generated by the baselines and \TITLE.
Compared with SAM3D, our architectural modifications and reinforcement learning reduce the inter-object collision ratio and floating distance by 51\% and 55\%, respectively, while also improving ADD-S.
Moreover, \TITLE achieves lower ADD-S, collision ratio, and floating distance than all other baselines.

We additionally report the results of applying PICASSO to SAM3D and \TITLE in \cref{tab:post_results}.
Although PICASSO improves the pose accuracy of both methods, it increases their collision ratios and substantially worsens the floating distance for \TITLE.
These results reveal trade-offs between pose accuracy and physical plausibility under PICASSO refinement.
Nevertheless, \TITLE combined with PICASSO outperforms SAM3D with PICASSO on four of the five metrics, suggesting that the initial reconstruction affects refinement outcomes.

\Cref{fig:results} complements the quantitative evaluation with qualitative comparisons.
In the illustrated scenes, \TITLE avoids inter-object collisions and maintains physically plausible contact relationships, whereas the baseline reconstructions exhibit collisions or floating artifacts.
Together, the qualitative and quantitative results also demonstrate synthetic-to-real transfer, as \TITLE achieves these improvements on real-world YCB-V~\citep{PoseCNN} inputs after fine-tuning on synthetic FoundationPose~\citep{FoundationPose} scenes.

\subsection{Ablation Study}

\subsubsection{Inter-Object Attention}

\Cref{tab:ablation} evaluates inter-object attention and positional encoding under our proposed RL method.
Applying RL alone (SAM3D + RL) reduces collision and floating relative to the pre-trained model, but worsens ADD-S and ground displacement.
Adding gated inter-object attention through simple concatenation without positional encoding degrades all reported metrics.
This suggests that enabling inter-object information exchange alone is insufficient to improve reconstruction.

\begin{table}[H]
\caption{Ablation Study on \textbf{Inter-Object Attention (IOA)} and \textbf{Positional Encoding.}}
\vspace{-2mm}
\label{tab:ablation}
\centering
\footnotesize
\setlength{\tabcolsep}{1pt}
\begin{tabularx}{\linewidth}{@{}l<{\hspace{4pt}}*{5}{>{\centering\arraybackslash}X}@{}}
\toprule
& \multicolumn{2}{c}{Alignment}
& \multicolumn{3}{c}{Physics} \\
\trimcline{2-3}\trimcline{4-6}
Variants
& ADD-S $\downarrow$ & \mbox{ADD-S@0.1$\!\uparrow$}
& \rule{0pt}{2.4ex} Collision $\downarrow$ & Floating $\downarrow$
& Ground $\downarrow$ \\
\midrule
SAM3D
& \second 0.0697 & \third 0.8962 & 10.24 & 3.35 \third & \second 9.47  \\
\quad + RL
& \third 0.0777 & \second 0.9469 & \third 5.40 & \first 1.29 & 14.73 \\
\quad + RL + IOA
& 0.2672 & 0.0480 & 13.10 & 12.43 & 33.90 \\
\quad + RL + IOA + Center-RoPE
& 0.0797 & 0.8781 & \first 4.23 & 7.95 & \first 9.40 \\
\quad \textbf{+ RL + IOA + Volume-RoPE}
& \first 0.0685 & \first 0.9556 & \second 5.01 & \second 1.50 & \third 11.20 \\
\bottomrule
\vspace{-5mm}
\end{tabularx}
\end{table}

Both Center-RoPE and Volume-RoPE improve performance over inter-object attention without positional encoding, supporting the importance of explicit spatial guidance.
Although Center-RoPE yields lower collision ratio and ground displacement than Volume-RoPE, it still produces higher ADD-S and more than twice the floating distance of the pre-trained SAM3D model.
By contrast, Volume-RoPE achieves the best alignment metrics among all evaluated variants while reducing both collision and floating relative to SAM3D.
These results demonstrate the benefit of incorporating object extents into positional encoding for accurate and physically plausible reconstruction.

\subsubsection{Object-Wise GDPO}

The quantitative results comparing Scene-level and Object-wise GDPO are reported in \cref{tab:ow_ablation}.

\begin{table}[H]
\caption{Ablation Study on \textbf{Object-wise GDPO.}}
\vspace{-4mm}
\label{tab:ow_ablation}
\begin{center}
\small
\begin{tabular*}{0.85\linewidth}{@{}lc@{\extracolsep{\fill}}cccc@{}}
    \toprule
     & \multicolumn{2}{c}{Alignment} & \multicolumn{3}{c}{Physics} \\
    \trimcline{2-3}\trimcline{4-6}
      & ADD-S $\downarrow$ & ADD-S@0.1 $\uparrow$ & \rule{0pt}{2.4ex}Collision $\downarrow$ & Floating $\downarrow$ & \rule{0pt}{2.4ex}Ground $\downarrow$ \\
    \midrule
    Scene-Level GDPO & \textbf{0.0605} & 0.9479 & 10.37 & \textbf{0.58} & 17.03  \\
    \textbf{Object-Wise GDPO} & 0.0685 & \textbf{0.9556} & \textbf{5.01} & 1.50 & \textbf{11.20} \\
    \bottomrule
\end{tabular*}
\end{center}
\vspace{-5mm}
\end{table}

With both approaches learning better GT alignment, Scene-level GDPO scored better in mean ADD-S and Object-wise GDPO in accuracy.
Also, both models prevented the floating of the pre-trained model, as demonstrated by the lower floating distance compared to SAM3D.
However, the collision ratio of Scene-level GDPO increased from SAM3D, while Object-wise GDPO was successfully trained to reduce collisions by half.
These results suggest that Scene-level GDPO is less effective at reducing inter-object collisions, consistent with our hypothesis in \cref{sec:ow_gdpo}.

\subsubsection{Physical-Plausibility Rewards}

We compare the full training objective with two variants in \cref{tab:physics_rl}.
The SFT-only variant minimizes the supervised loss $\Ls_{\mathrm{SFT}}$ from \cref{sec:sft}, while the GT-reward-only variant uses the ground-truth pose reward in \cref{eq:gt_reward} as its sole RL reward.
Because each variant takes less time per training step, we train both for 750 steps, approximately matching the training time.

\begin{table}[H]
\caption{Ablation Study on \textbf{Physical-Plausibility Rewards.}}
\vspace{-2mm}
\label{tab:physics_rl}
\centering
\footnotesize
\setlength{\tabcolsep}{1pt}
\begin{tabularx}{\linewidth}{@{}l<{\hspace{4pt}}*{5}{>{\centering\arraybackslash}X}@{}}
\toprule
& \multicolumn{2}{c}{Alignment} & \multicolumn{3}{c}{Physics} \\
\trimcline{2-3}\trimcline{4-6}
Variants & ADD-S $\downarrow$ & \mbox{ADD-S@0.1$\!\uparrow$} & \rule{0pt}{2.4ex} Collision $\downarrow$ & Floating $\downarrow$ & Ground $\downarrow$ \\
\midrule
SFT only & \second 0.0690 & 0.9278 & 10.69 & \first 1.08 & 11.86 \\
GT reward only & 0.0704 & \second 0.9423 & \second 8.63 & 6.63 & \first 9.05 \\
\textbf{Full objective} & \first 0.0685 & \first 0.9556 & \first 5.01 & \second 1.50 & \second 11.20 \\
\bottomrule
\vspace{-7mm}
\end{tabularx}
\end{table}

Among the three training objectives, the full objective achieves the best ADD-S, ADD-S@0.1, and collision ratio and ranks second on floating distance and ground displacement.
Relative to the pre-trained SAM3D model in \cref{tab:results}, SFT alone increases the collision ratio, while using only the GT reward increases both ADD-S and floating distance.
The results suggest that reinforcement learning with physical-plausibility rewards improves both reconstruction quality and physical plausibility.

\subsection{Analysis}

\begin{wrapfigure}{r}{0.4\textwidth}
    \vspace{-6mm}
    \centering
    \includegraphics[width=\linewidth]{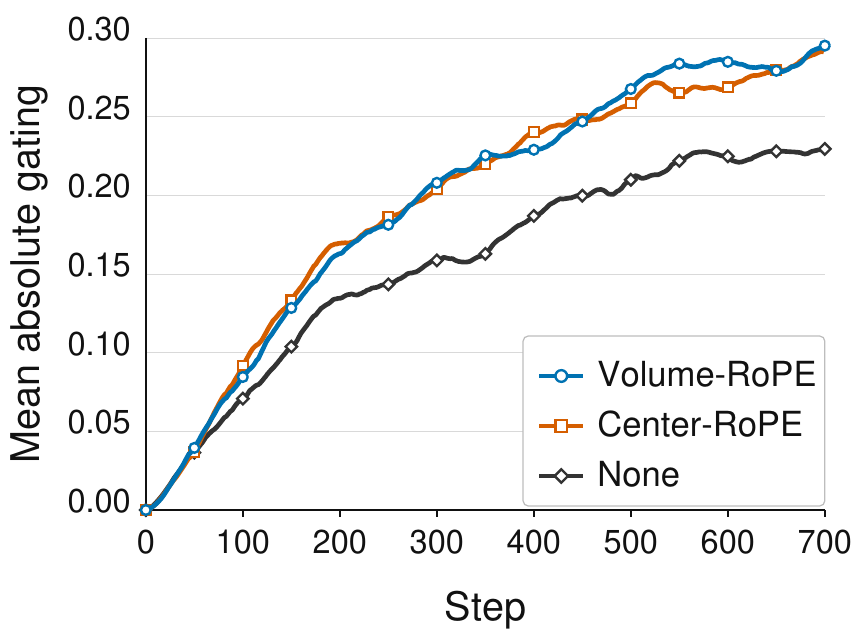}
    \vspace{-6mm}
    \caption{Evolution of the Gating Parameter $g$}
    \label{fig:gating}
\end{wrapfigure}

\textbf{Gating Factor.}
The trainable gating parameters $g$ control the contribution of inter-object attention to pose estimation.
\Cref{fig:gating} plots the mean absolute value of $g$ across layers and attention heads over the course of training.
As expected, the increasing magnitude of $g$ indicates that the model assigns progressively greater weight to inter-object attention during training.
By step 500, the mean magnitude reaches approximately 0.25 and increases further to 0.30 by step 700.
Volume-RoPE and Center-RoPE exhibit similar gating dynamics.
Without RoPE, however, the magnitude of $g$ grows more slowly than with the RoPE-based variants.
This slower growth suggests that learning inter-object interactions is more difficult in the absence of positional encoding.

\textbf{Attention Map.} Volume-RoPE prompts inter-object attention to attend more on physically adjacent objects according to its attention map, compared to the inter-object attention without RoPE. The detailed analysis is in \cref{sec:attention_map}

\section{Conclusion}

\textbf{Conclusion.}
In this paper, we presented \TITLE, a unified architecture and training framework for physically plausible 3D scene reconstruction with a joint shape-and-pose generator.
Our \textit{Gated Inter-Object Attention} models relationships among objects while faithfully preserving the pre-trained model's intra-object attention behavior at initialization, and \ontest{\textit{Volume-RoPE}} provides explicit spatial information.
By reinforcing the pre-trained model with \textit{Object-wise GDPO} and \textit{Physical-Plausibility Rewards}, \TITLE outperforms the evaluated baselines on most metrics for reconstruction accuracy and physical plausibility.
We further validated the effectiveness of each proposed architectural and training component through extensive experiments.

\textbf{Limitation.}
Although Object-wise GDPO is broadly applicable to various reward functions, we focus only on object collisions and floating as an initial step toward physically plausible 3D reconstruction.
Future work could extend the framework to a broader range of physical properties, such as stability, by incorporating additional reward functions.
Moreover, although our method generally achieves better physical plausibility than previous approaches, it cannot \textit{guarantee} physically plausible outputs in every case, which remains an important consideration for real-world applications.
Finally, the $\sinc(\freq h_i)$ term causes Volume-RoPE to violate an important property of RoPE: norm preservation.
Standard RoPE rotates the query and key while preserving their norms, whereas Volume-RoPE does not.
Although we observed no issues in our setting, this violation may cause problems when Volume-RoPE is combined with techniques that rely on norm preservation.

\newpage

\bibliography{iclr2027_conference}

\begin{thebibliography}{39}
\providecommand{\natexlab}[1]{#1}
\providecommand{\url}[1]{\texttt{#1}}
\expandafter\ifx\csname urlstyle\endcsname\relax
  \providecommand{\doi}[1]{doi: #1}\else
  \providecommand{\doi}{doi: \begingroup \urlstyle{rm}\Url}\fi

\bibitem[Blattmann et~al.(2023)Blattmann, Rombach, Ling, Dockhorn, Kim, Fidler, and Kreis]{Video-LDM}
Andreas Blattmann, Robin Rombach, Huan Ling, Tim Dockhorn, Seung~Wook Kim, Sanja Fidler, and Karsten Kreis.
\newblock Align your latents: High-resolution video synthesis with latent diffusion models.
\newblock In \emph{Conference on Computer Vision and Pattern Recognition}, 2023.

\bibitem[Chen et~al.(2026)Chen, Chu, Gleize, Liang, Sax, Tang, Wang, Guo, Hardin, Li, et~al.]{SAM3D}
Xingyu Chen, Fu-Jen Chu, Pierre Gleize, Kevin~J Liang, Alexander Sax, Hao Tang, Weiyao Wang, Michelle Guo, Thibaut Hardin, Xiang Li, et~al.
\newblock Sam 3d: 3dfy anything in images.
\newblock In \emph{Conference on Computer Vision and Pattern Recognition}, 2026.

\bibitem[Dong et~al.(2026)Dong, Deng, Zhang, Dai, Deng, and Tang]{CAWorld}
Xin Dong, Weijian Deng, Lihan Zhang, Tianru Dai, Wenfeng Deng, and Yansong Tang.
\newblock Ca-world: Multi-object counterfactual alignment for efficient interactive-ready reconstruction.
\newblock \emph{arXiv preprint arXiv:2605.30239}, 2026.

\bibitem[Downs et~al.(2022)Downs, Francis, Koenig, Kinman, Hickman, Reymann, McHugh, and Vanhoucke]{GSO}
Laura Downs, Anthony Francis, Nate Koenig, Brandon Kinman, Ryan Hickman, Krista Reymann, Thomas~B McHugh, and Vincent Vanhoucke.
\newblock Google scanned objects: A high-quality dataset of 3d scanned household items.
\newblock In \emph{International Conference on Robotics and Automation}, 2022.

\bibitem[Ho et~al.(2020)Ho, Jain, and Abbeel]{DDPM}
Jonathan Ho, Ajay Jain, and Pieter Abbeel.
\newblock Denoising diffusion probabilistic models.
\newblock In \emph{Advances in Neural Information Processing Systems}, 2020.

\bibitem[Hu et~al.(2022)Hu, Shen, Wallis, Allen-Zhu, Li, Wang, Wang, and Chen]{LoRA}
Edward~J Hu, Yelong Shen, Phillip Wallis, Zeyuan Allen-Zhu, Yuanzhi Li, Shean Wang, Lu~Wang, and Weizhu Chen.
\newblock Lo{RA}: Low-rank adaptation of large language models.
\newblock In \emph{International Conference on Learning Representations}, 2022.

\bibitem[Huang et~al.(2026{\natexlab{a}})Huang, Duan, Zhao, Zhao, Ma, and Gao]{CUPID}
Binbin Huang, Haobin Duan, Yiqun Zhao, Zibo Zhao, Yi~Ma, and Shenghua Gao.
\newblock Cupid: Generative 3d reconstruction via joint object and pose modeling.
\newblock \emph{Conference on Computer Vision and Pattern Recognition}, 2026{\natexlab{a}}.

\bibitem[Huang et~al.(2026{\natexlab{b}})Huang, Han, Ye, Pan, and Hauser]{SPARCS}
Wei-Cheng Huang, Jiaheng Han, Xiaohan Ye, Zherong Pan, and Kris Hauser.
\newblock Simulation-ready cluttered scene estimation via physics-aware joint shape and pose optimization.
\newblock \emph{Robotics: Science and Systems}, 2026{\natexlab{b}}.

\bibitem[Huang et~al.(2026{\natexlab{c}})Huang, Li, He, Zhou, and Shechtman]{SelfForcing}
Xun Huang, Zhengqi Li, Guande He, Mingyuan Zhou, and Eli Shechtman.
\newblock Self forcing: Bridging the train-test gap in autoregressive video diffusion.
\newblock 2026{\natexlab{c}}.

\bibitem[Huang et~al.(2025)Huang, Guo, An, Yang, Li, Zou, Liang, Liu, Cao, and Sheng]{MIDI}
Zehuan Huang, Yuan-Chen Guo, Xingqiao An, Yunhan Yang, Yangguang Li, Zi-Xin Zou, Ding Liang, Xihui Liu, Yan-Pei Cao, and Lu~Sheng.
\newblock Midi: Multi-instance diffusion for single image to 3d scene generation.
\newblock In \emph{Conference on Computer Vision and Pattern Recognition}, 2025.

\bibitem[Labb\'e et~al.(2022)Labb\'e, Manuelli, Mousavian, Tyree, Birchfield, Tremblay, Carpentier, Aubry, Fox, and Sivic]{MegaPose}
Yann Labb\'e, Lucas Manuelli, Arsalan Mousavian, Stephen Tyree, Stan Birchfield, Jonathan Tremblay, Justin Carpentier, Mathieu Aubry, Dieter Fox, and Josef Sivic.
\newblock Megapose: 6d pose estimation of novel objects via render \& compare.
\newblock In \emph{Conference on Robot Learning}, 2022.

\bibitem[Lee et~al.(2026)Lee, Baik, Kim, Kim, Cha, and Joo]{SimuScene}
Inhee Lee, Sangwon Baik, Sungjoo Kim, Hyeonwoo Kim, Hyunsoo Cha, and Hanbyul Joo.
\newblock Simuscene: Simulation-ready compositional 3d scene reconstruction from a single image.
\newblock \emph{arXiv preprint arXiv:2606.03994}, 2026.

\bibitem[Li et~al.(2026{\natexlab{a}})Li, Wu, Li, Zhou, Zeng, Li, and Zha]{MV-SAM3D}
Baicheng Li, Dong Wu, Jun Li, Shunkai Zhou, Zecui Zeng, Lusong Li, and Hongbin Zha.
\newblock Mv-sam3d: Adaptive multi-view fusion for layout-aware 3d generation.
\newblock \emph{arXiv preprint arXiv:2603.11633}, 2026{\natexlab{a}}.

\bibitem[Li et~al.(2026{\natexlab{b}})Li, Shao, Lu, Fu, Chen, Jain, and Chandraker]{PhiScene}
Haodong Li, Lulu Shao, Haolin Lu, Yu~Fu, Yen-Ru Chen, Seemandhar Jain, and Manmohan Chandraker.
\newblock $\phi$-scene: Physically grounded image-to-3d scene reconstruction.
\newblock \emph{arXiv preprint arXiv:2606.21596}, 2026{\natexlab{b}}.

\bibitem[Li et~al.(2025{\natexlab{a}})Li, Zheng, Rupprecht, and Vedaldi]{DSO}
Ruining Li, Chuanxia Zheng, Christian Rupprecht, and Andrea Vedaldi.
\newblock Dso: Aligning 3d generators with simulation feedback for physical soundness.
\newblock In \emph{International Conference on Computer Vision}, 2025{\natexlab{a}}.

\bibitem[Li et~al.(2025{\natexlab{b}})Li, Zou, Liu, Wang, Liang, Yu, Liu, Guo, Liang, Ouyang, et~al.]{TripoSG}
Yangguang Li, Zi-Xin Zou, Zexiang Liu, Dehu Wang, Yuan Liang, Zhipeng Yu, Xingchao Liu, Yuan-Chen Guo, Ding Liang, Wanli Ouyang, et~al.
\newblock Triposg: High-fidelity 3d shape synthesis using large-scale rectified flow models.
\newblock \emph{arXiv preprint arXiv:2502.06608}, 2025{\natexlab{b}}.

\bibitem[Lin et~al.(2024)Lin, Liu, Lu, and Jia]{SAM6D}
Jiehong Lin, Lihua Liu, Dekun Lu, and Kui Jia.
\newblock Sam-6d: Segment anything model meets zero-shot 6d object pose estimation.
\newblock In \emph{Conference on Computer Vision and Pattern Recognition}, 2024.

\bibitem[Lipman et~al.(2023)Lipman, Chen, Ben-Hamu, Nickel, and Le]{Flow}
Yaron Lipman, Ricky T.~Q. Chen, Heli Ben-Hamu, Maximilian Nickel, and Matt Le.
\newblock Flow matching for generative modeling.
\newblock In \emph{International Conference on Learning Representations}, 2023.

\bibitem[Liu et~al.(2025)Liu, Liu, Liang, Li, Liu, Wang, Wan, Zhang, and Ouyang]{FlowGRPO}
Jie Liu, Gongye Liu, Jiajun Liang, Yangguang Li, Jiaheng Liu, Xintao Wang, Pengfei Wan, Di~Zhang, and Wanli Ouyang.
\newblock Flow-grpo: Training flow matching models via online rl.
\newblock In \emph{Advances in Neural Information Processing Systems}, 2025.

\bibitem[Liu et~al.(2026)Liu, Dong, Lu, Diao, Belcak, Liu, Chen, Yin, Wang, Cheng, et~al.]{GDPO}
Shih-Yang Liu, Xin Dong, Ximing Lu, Shizhe Diao, Peter Belcak, Mingjie Liu, Min-Hung Chen, Hongxu Yin, Yu-Chiang~Frank Wang, Kwang-Ting Cheng, et~al.
\newblock Gdpo: Group reward-decoupled normalization policy optimization for multi-reward rl optimization.
\newblock In \emph{International Conference on Machine Learning}, 2026.

\bibitem[Ma et~al.(2026)Ma, Wang, Ugrinovic, Litman, and Kitani]{REST3D}
Xiaoxuan Ma, Jiashun Wang, Nicol\'{a}s Ugrinovic, Yehonathan Litman, and Kris Kitani.
\newblock {REST3D}: Reconstructing physically stable 3d scenes from a single image.
\newblock 2026.

\bibitem[Meng et~al.(2026)Meng, Wu, Zhang, and Xie]{SceneGen}
Yanxu Meng, Haoning Wu, Ya~Zhang, and Weidi Xie.
\newblock Scenegen: Single-image 3d scene generation in one feedforward pass.
\newblock In \emph{International Conference on 3D Vision}, 2026.

\bibitem[Ouyang et~al.(2022)Ouyang, Wu, Jiang, Almeida, Wainwright, Mishkin, Zhang, Agarwal, Slama, Ray, et~al.]{RLHF}
Long Ouyang, Jeffrey Wu, Xu~Jiang, Diogo Almeida, Carroll Wainwright, Pamela Mishkin, Chong Zhang, Sandhini Agarwal, Katarina Slama, Alex Ray, et~al.
\newblock Training language models to follow instructions with human feedback.
\newblock \emph{Advances in Neural Information Processing Systems}, 2022.

\bibitem[Rafailov et~al.(2023)Rafailov, Sharma, Mitchell, Manning, Ermon, and Finn]{DPO}
Rafael Rafailov, Archit Sharma, Eric Mitchell, Christopher~D Manning, Stefano Ermon, and Chelsea Finn.
\newblock Direct preference optimization: Your language model is secretly a reward model.
\newblock In \emph{Advances in Neural Information Processing Systems}, 2023.

\bibitem[Ross et~al.(2011)Ross, Gordon, and Bagnell]{DAgger}
St{\'e}phane Ross, Geoffrey Gordon, and Drew Bagnell.
\newblock A reduction of imitation learning and structured prediction to no-regret online learning.
\newblock In \emph{Proceedings of the fourteenth international conference on artificial intelligence and statistics}, pp.\  627--635. JMLR Workshop and Conference Proceedings, 2011.

\bibitem[Shao et~al.(2024)Shao, Wang, Zhu, Xu, Song, Bi, Zhang, Zhang, Li, Wu, et~al.]{DeepSeekMath}
Zhihong Shao, Peiyi Wang, Qihao Zhu, Runxin Xu, Junxiao Song, Xiao Bi, Haowei Zhang, Mingchuan Zhang, YK~Li, Yang Wu, et~al.
\newblock Deepseekmath: Pushing the limits of mathematical reasoning in open language models.
\newblock \emph{arXiv preprint arXiv:2402.03300}, 2024.

\bibitem[Su et~al.(2024)Su, Ahmed, Lu, Pan, Bo, and Liu]{Roformer}
Jianlin Su, Murtadha Ahmed, Yu~Lu, Shengfeng Pan, Wen Bo, and Yunfeng Liu.
\newblock Roformer: Enhanced transformer with rotary position embedding.
\newblock \emph{Neurocomputing}, 568:\penalty0 127063, 2024.

\bibitem[Wallace et~al.(2024)Wallace, Dang, Rafailov, Zhou, Lou, Purushwalkam, Ermon, Xiong, Joty, and Naik]{DiffusionDPO}
Bram Wallace, Meihua Dang, Rafael Rafailov, Linqi Zhou, Aaron Lou, Senthil Purushwalkam, Stefano Ermon, Caiming Xiong, Shafiq Joty, and Nikhil Naik.
\newblock Diffusion model alignment using direct preference optimization.
\newblock In \emph{Conference on Computer Vision and Pattern Recognition}, 2024.

\bibitem[Wen et~al.(2024)Wen, Yang, Kautz, and Birchfield]{FoundationPose}
Bowen Wen, Wei Yang, Jan Kautz, and Stan Birchfield.
\newblock Foundationpose: Unified 6d pose estimation and tracking of novel objects.
\newblock In \emph{Conference on Computer Vision and Pattern Recognition}, 2024.

\bibitem[Wu et~al.(2026)Wu, Hu, Hui, Wei, Luo, Li, and Liu]{PhyMix}
Dongli Wu, Jingyu Hu, Ka-Hei Hui, Xiaobao Wei, Chengwen Luo, Jianqiang Li, and Zhengzhe Liu.
\newblock Phymix: Towards physically consistent single-image 3d indoor scene generation with implicit--explicit optimization.
\newblock \emph{arXiv preprint arXiv:2604.10125}, 2026.

\bibitem[Wu et~al.(2025)Wu, Lin, Zhang, Zeng, Yang, Bao, Qian, Zhu, Torr, Cao, and Yao]{Direct3DS2}
Shuang Wu, Youtian Lin, Feihu Zhang, Yifei Zeng, Yikang Yang, Yajie Bao, Jiachen Qian, Siyu Zhu, Philip Torr, Xun Cao, and Yao Yao.
\newblock Direct3d-s2: Gigascale 3d generation made easy with spatial sparse attention.
\newblock \emph{arXiv preprint arXiv:2505.17412}, 2025.

\bibitem[Xiang et~al.(2025)Xiang, Lv, Xu, Deng, Wang, Zhang, Chen, Tong, and Yang]{TRELLIS}
Jianfeng Xiang, Zelong Lv, Sicheng Xu, Yu~Deng, Ruicheng Wang, Bowen Zhang, Dong Chen, Xin Tong, and Jiaolong Yang.
\newblock Structured 3d latents for scalable and versatile 3d generation.
\newblock In \emph{Conference on Computer Vision and Pattern Recognition}, 2025.

\bibitem[Xiang et~al.(2026)Xiang, Chen, Xu, Wang, Lv, Deng, Zhu, Dong, Zhao, Yuan, et~al.]{TRELLIS2}
Jianfeng Xiang, Xiaoxue Chen, Sicheng Xu, Ruicheng Wang, Zelong Lv, Yu~Deng, Hongyuan Zhu, Yue Dong, Hao Zhao, Nicholas~Jing Yuan, et~al.
\newblock Native and compact structured latents for 3d generation.
\newblock In \emph{Conference on Computer Vision and Pattern Recognition}, 2026.

\bibitem[Xiang et~al.(2017)Xiang, Schmidt, Narayanan, and Fox]{PoseCNN}
Yu~Xiang, Tanner Schmidt, Venkatraman Narayanan, and Dieter Fox.
\newblock Posecnn: A convolutional neural network for 6d object pose estimation in cluttered scenes.
\newblock \emph{Robotics: Science and Systems}, 2017.

\bibitem[Yang et~al.(2024)Yang, Shi, Zhang, Yang, Wang, Zhao, Liu, Wang, Lin, Yu, Wang, Xu, He, Chen, Liu, Wu, Lian, Yang, Liu, Yang, Wang, Jiang, and Guo]{Hunyuan3D}
Xianghui Yang, Huiwen Shi, Bowen Zhang, Fan Yang, Jiacheng Wang, Hongxu Zhao, Xinhai Liu, Xinzhou Wang, Qingxiang Lin, Jiaao Yu, Lifu Wang, Jing Xu, Zebin He, Zhuo Chen, Sicong Liu, Junta Wu, Yihang Lian, Shaoxiong Yang, Yuhong Liu, Yong Yang, Di~Wang, Jie Jiang, and Chunchao Guo.
\newblock {Hunyuan3D} 1.0: A unified framework for text-to-{3D} and image-to-{3D} generation.
\newblock \emph{arXiv preprint arXiv:2411.02293}, 2024.

\bibitem[Yu et~al.(2026)Yu, Talak, Shaikewitz, and Carlone]{PICASSO}
Xihang Yu, Rajat Talak, Lorenzo Shaikewitz, and Luca Carlone.
\newblock Picasso: Holistic scene reconstruction with physics-constrained sampling.
\newblock \emph{Robotics: Science and Systems}, 2026.

\bibitem[Zadaianchuk et~al.(2026)Zadaianchuk, Barcellona, Schuenemann, Gumbsch, Wang, Irshad, Despinoy, Aljundi, Gavves, and Zakharov]{RecGen}
Andrii Zadaianchuk, Leonardo Barcellona, Lennard Schuenemann, Christian Gumbsch, Zehao Wang, Muhammad~Zubair Irshad, Fabien Despinoy, Rahaf Aljundi, Stratis Gavves, and Sergey Zakharov.
\newblock Reconstruction by generation: 3d multi-object scene reconstruction from sparse observations.
\newblock In \emph{European Conference on Computer Vision}, 2026.

\bibitem[Zhang et~al.(2023)Zhang, Tang, Niessner, and Wonka]{VecSet}
Biao Zhang, Jiapeng Tang, Matthias Niessner, and Peter Wonka.
\newblock 3dshape2vecset: A 3d shape representation for neural fields and generative diffusion models.
\newblock \emph{ACM Transactions on Graphics}, 42\penalty0 (4):\penalty0 1--16, 2023.

\bibitem[Zhang et~al.(2024)Zhang, Han, Liu, Gao, Zhou, Hu, Yan, Lu, Li, and Qiao]{LLaMA-Adapter}
Renrui Zhang, Jiaming Han, Chris Liu, Peng Gao, Aojun Zhou, Xiangfei Hu, Shilin Yan, Pan Lu, Hongsheng Li, and Yu~Qiao.
\newblock Llama-adapter: Efficient fine-tuning of language models with zero-init attention.
\newblock In \emph{International Conference on Learning Representations}, 2024.

\end{thebibliography}
\bibliographystyle{iclr2027_conference}

\newpage

\appendix
\section{Appendix}

\subsection{Volume-RoPE: Derivation and Implementation}
\label{sec:rope_math}
In this section, mathematical details of \cref{sec:volume_rope} is explained.

The equation below is a derivation of \cref{eq:volume_rope}. Using the substitutions $u=x-t_i$ and $v=y-t_j$, we obtain
\begin{align}
&\frac{1}{4h_ih_j}\int_{t_j-h_j}^{t_j+h_j}\int_{t_i-h_i}^{t_i+h_i}e^{i\freq(x-y)}\mathrm{d}x\mathrm{d}y \nonumber \\
&=\frac{1}{4h_ih_j}\left(\int_{t_i-h_i}^{t_i+h_i}e^{i\freq x}\mathrm{d}x\right)\left(\int_{t_j-h_j}^{t_j+h_j}e^{-i\freq y}\mathrm{d}y\right) \nonumber \\
&=\frac{e^{i\freq(t_i-t_j)}}{4h_ih_j}\left(\int_{-h_i}^{h_i}e^{i\freq u}\mathrm{d}u\right)\left(\int_{-h_j}^{h_j}e^{-i\freq v}\mathrm{d}v\right) \nonumber \\
&=\frac{e^{i\freq(t_i-t_j)}}{4h_ih_j}
\left(\frac{e^{i\freq h_i}-e^{-i\freq h_i}}{i\freq}\right)
\left(\frac{e^{-i\freq h_j}-e^{i\freq h_j}}{-i\freq}\right) \nonumber \\
&=\frac{e^{i\freq(t_i-t_j)}}{4h_ih_j}
\left(\frac{2\sin(\freq h_i)}{\freq}\right)
\left(\frac{2\sin(\freq h_j)}{\freq}\right) \nonumber \\
&=\frac{\sin(\freq h_i)}{\freq h_i}\frac{\sin(\freq h_j)}{\freq h_j}e^{i\freq(t_i-t_j)} \nonumber \\
&=\sinc(\freq h_i)\sinc(\freq h_j)e^{i\freq(t_i-t_j)},
\end{align}
where $\sinc(z)=\sin(z)/z$ for $z\neq 0$ and $\sinc(0)=1$ by continuity.
Consequently, the mean relative embedding factorizes as
\begin{equation}
\left(\sinc(\freq h_i)e^{i\freq t_i}\right)
\overline{\left(\sinc(\freq h_j)e^{i\freq t_j}\right)},
\end{equation}
which yields the Volume-RoPE embedding $\sinc(\freq h_i)e^{i\freq t_i}$ for object $i$.

For the practical formulation described in \cref{sec:volume_rope}, we replace the translation $t_i$ with the bounding-box center $c_i$ estimated from the masked pointmap.
We then standardize $\nset{c}$ and $\nset{h}$ using the mean and standard deviation of $\nset{c}$; hereafter, $c_i$ and $h_i$ denote these standardized quantities.

To apply Volume-RoPE to real-valued queries and keys, we represent each complex multiplication using a two-dimensional block.
Let $d=6M$ be the query and key dimension, where $M$ is the number of frequencies assigned to each spatial axis.
We first define the two-dimensional rotation matrix
\begin{equation}
\mR_{\mathrm{2D}}(\theta)=
\begin{bmatrix}
\cos\theta & -\sin\theta \\
\sin\theta & \cos\theta
\end{bmatrix}.
\end{equation}
For frequency $\freq_m$ and axis $a\in\{x,y,z\}$, the Volume-RoPE block for object $i$ is
\begin{equation}
\label{eq:volume_rope_block}
\mV_{i,m,a}
=\sinc(\freq_m h_i^a)\mR_{\mathrm{2D}}(\freq_m c_i^a).
\end{equation}
Stacking these blocks in the order $(1,x),(1,y),(1,z),\ldots,(M,x),(M,y),(M,z)$ gives
\begin{equation}
\label{eq:volume_rope_matrix}
\mV_i
=\bigoplus_{m=1}^{M}\bigoplus_{a\in\{x,y,z\}}\mV_{i,m,a},
\end{equation}
where $\bigoplus$ denotes block-diagonal concatenation.
Volume-RoPE is then applied to the query and key vectors as
\begin{equation}
\label{eq:volume_rope_qk}
\widetilde{\vq}_i=\mV_i\vq_i,
\qquad
\widetilde{\vk}_i=\mV_i\vk_i.
\end{equation}

Because $\mV_i$ is block diagonal, \cref{eq:volume_rope_qk} acts independently on each frequency--axis block: $\widetilde{\vq}_{i,m,a}=\mV_{i,m,a}\vq_{i,m,a}$ and $\widetilde{\vk}_{j,m,a}=\mV_{j,m,a}\vk_{j,m,a}$.
Substituting these definitions into the dot product and using $(\mA\vx)^{\top}=\vx^{\top}\mA^{\top}$ gives
\begin{equation}
\label{eq:volume_rope_query_key_product}
\begin{aligned}
\widetilde{\vq}_{i,m,a}^{\top}\widetilde{\vk}_{j,m,a}
&=\left(\mV_{i,m,a}\vq_{i,m,a}\right)^{\top}
\left(\mV_{j,m,a}\vk_{j,m,a}\right) \\
&=\vq_{i,m,a}^{\top}\mV_{i,m,a}^{\top}\mV_{j,m,a}\vk_{j,m,a},
\end{aligned}
\end{equation}
The matrix $\mV_{i,m,a}^{\top}\mV_{j,m,a}$ in \cref{eq:volume_rope_query_key_product} has a direct interpretation.
Substituting \cref{eq:volume_rope_block} and using $\mR_{\mathrm{2D}}(\alpha)^{\top}\mR_{\mathrm{2D}}(\beta)=\mR_{\mathrm{2D}}(\beta-\alpha)$ gives
\begin{equation}
\label{eq:volume_rope_relative_matrix}
\mV_{i,m,a}^{\top}\mV_{j,m,a}
=\sinc(\freq_m h_i^a)\sinc(\freq_m h_j^a)
\mR_{\mathrm{2D}}\left(\freq_m(c_j^a-c_i^a)\right).
\end{equation}
Thus, \cref{eq:volume_rope_query_key_product} is a standard relative-position RoPE interaction weighted by the product of the two objects' extent-dependent $\sinc$ terms.
Under the column-vector convention, the relative rotation $\freq_m(c_j^a-c_i^a)$ corresponds to the complex phase $e^{i\freq_m(c_i^a-c_j^a)}$ in \cref{eq:volume_rope}.

\subsection{Stochastic Flow Policy and KL Regularization}
\label{sec:rl}

We review how the deterministic flow sampler defines stochastic $\pitheta$ and why $\KL(\pitheta\|\piref)$ admits a closed form, following Flow-GRPO~\citep{FlowGRPO}.
The original flow sampler is deterministic, so Flow-GRPO converts it into an equivalent reverse-time SDE to obtain a stochastic policy for RL.
Applying Euler--Maruyama discretization to this SDE gives
\begin{equation}
\label{eq:flow_grpo_transition}
\vx_{t+\Delta t}
=\vx_t
+\left[
\vv_\theta(\vx_t,t,\mathcal{C})
+\frac{\sigma_t^2}{2t}
\left(\vx_t+(1-t)\vv_\theta(\vx_t,t,\mathcal{C})\right)
\right]\Delta t
+\sigma_t\sqrt{\Delta t}\,\boldsymbol{\epsilon},
\qquad
\boldsymbol{\epsilon}\sim\mathcal{N}(\mathbf{0},\mathbf{I}),
\end{equation}
where $\vv_\theta$ is the velocity predicted by the trainable flow model, $\Delta t$ is the discretization step, and $\mathcal{C}$ denotes the conditioning inputs.
Following Flow-GRPO, we set the noise schedule to
\begin{equation}
\sigma_t=\alpha\sqrt{\frac{t}{1-t}},
\end{equation}
where the scalar hyperparameter $\alpha$ controls the amount of stochasticity during sampling.
We set $\alpha=0.3$ in our experiments.
Conditioned on $\vx_t$ and $\mathcal{C}$, the only random term in \cref{eq:flow_grpo_transition} is $\sigma_t\sqrt{\Delta t}\,\boldsymbol{\epsilon}$.
Therefore, denoting all remaining terms by $\bar{\vx}_{t+\Delta t,\theta}$, the one-step policy is the isotropic Gaussian
\begin{equation}
\pitheta(\vx_{t+\Delta t}\mid\vx_t,\mathcal{C})
=\mathcal{N}\!\left(
\bar{\vx}_{t+\Delta t,\theta},
\sigma_t^2\Delta t\,\mathbf{I}
\right).
\end{equation}
The reference policy replaces $\vv_\theta$ with $\vv_\mathrm{ref}$ while retaining the same covariance.
For two Gaussian distributions with the same covariance, the KL divergence depends only on the difference between their means.
In this case, it has the closed form
\begin{align}
\KL(\pitheta\|\piref)
&=\frac{
\left\|\bar{\vx}_{t+\Delta t,\theta}
-\bar{\vx}_{t+\Delta t,\mathrm{ref}}\right\|_2^2
}{2\sigma_t^2\Delta t} \\
&=\frac{\Delta t}{2}
\left(
\frac{\sigma_t(1-t)}{2t}
+\frac{1}{\sigma_t}
\right)^2
\left\|
\vv_\theta(\vx_t,t,\mathcal{C})
-\vv_\mathrm{ref}(\vx_t,t,\mathcal{C})
\right\|_2^2.
\end{align}

\subsection{Trajectory-Based Supervised Fine-Tuning}
\label{sec:sft}

This section details the formulation of $\Ls_\mathrm{SFT}(\theta)$ in \cref{eq:loss}.
Given the ground-truth poses $\vx_\mathrm{gt}$ for a scene, the flow-matching objective~\citep{Flow} trains the model by minimizing the following velocity-prediction error:
\begin{equation}
\E_{\substack{
t\sim\mathcal{U}(0,1),\,
\boldsymbol{\epsilon}\sim\mathcal{N}(\mathbf{0},\mathbf{I})
}}
\left[
\left\|
\vv_\theta\left((1-t)\vx_\mathrm{gt}+t\boldsymbol{\epsilon},t,\mathcal{C}\right)
-\left(\boldsymbol{\epsilon}-\vx_\mathrm{gt}\right)
\right\|_2^2
\right].
\end{equation}
Rather than training $v_\theta$ on $\left((1-t)\vx_\mathrm{gt}+t\boldsymbol{\epsilon},t,\mathcal{C}\right)$, we train it on $(\vx_i^t, \frac{t}{T}, \mathcal{C})$, where each $\vx_i$ is sampled as in \cref{eq:objective}.
This approach offers two benefits.
First, the $\vv_\theta$ computations performed during reinforcement learning can be reused, improving computational efficiency.
Second, training on the sampled trajectories makes the model more robust to error accumulation and trajectory shifts induced by reinforcement learning.
This rationale parallels those of DAgger~\citep{DAgger} in imitation learning or Self Forcing~\citep{SelfForcing} in video diffusion.
For a sampled state $\vx_i^t$, whose normalized time is $\frac{t}{T}$, the target velocity is $\frac{T}{t}(\vx_i^t-\vx_\mathrm{gt})$, which reaches $\vx_\mathrm{gt}$ when integrated backward to time $0$.
Consequently, averaging over the $G$ sampled trajectories and their $T$ predefined time steps, the SFT loss in \cref{eq:loss} is given by
\begin{equation}
\begin{aligned}
\Ls_\mathrm{SFT}(\theta)
&=\frac{1}{GT}\sum_{i=1}^{G}\sum_{t=1}^{T}
\left\|
\vv_\theta\left(\vx_i^t,\frac{t}{T},\mathcal{C}\right)
-\frac{T}{t}\left(\vx_i^t-\vx_\mathrm{gt}\right)
\right\|_2^2.
\end{aligned}
\end{equation}

\clearpage
\subsection{Physical Stability in Simulation}
\label{sec:simulation_stability}

We evaluate physical stability on YCB-V using simulation-based metrics similar to those used in REST3D~\citep{REST3D}.
We classify an object as stable if its translational drift is below 5\,cm and its rotational drift is below 0.3\,rad.

\begin{table}[!htbp]
\caption{\textbf{Physical Stability.} Quantitative comparison on YCB-V.}
\vspace{-2mm}
\label{tab:simulation_stability}
\centering
\footnotesize
\setlength{\tabcolsep}{1pt}
\begin{tabularx}{\linewidth}{@{}l<{\hspace{4pt}}*{3}{>{\centering\arraybackslash}X}c@{}}
\toprule
Method & \mbox{Stable~(\%) $\uparrow$} & \mbox{Drift~(mm) $\downarrow$} & \mbox{Lin. vel.~(m/s) $\downarrow$} & \mbox{Ang. vel.~(rad/s) $\downarrow$} \\
\midrule
MIDI~\citep{MIDI} + ICP & 47.35 & 53.33 & 0.4033 & 8.7947 \\
RecGen~\citep{RecGen} & \third 81.99 & \third 24.96 & \third 0.2151 & \third 7.1263 \\
SAM3D~\citep{SAM3D} & \second 89.60 & \second 21.57 & \second 0.1997 & \second 4.8212 \\
\midrule
\textbf{\TITLE (Ours)} & \first 91.01 & \first 19.04 & \first 0.1637 & \first 3.8592 \\
\bottomrule
\end{tabularx}
\end{table}

Although we do not explicitly enforce physical stability during training, \TITLE outperforms the baselines on all four simulation metrics in \cref{tab:simulation_stability}.
We attribute these gains to the lower collision ratio and floating distance in \cref{tab:results}, since collisions and floating objects can reduce physical stability.

\begin{table}[!htbp]
\caption{Ablation Study on \textbf{Inter-Object Attention (IOA)} and \textbf{Positional Encoding} in Simulation.}
\vspace{-2mm}
\label{tab:stability_ioa}
\centering
\footnotesize
\setlength{\tabcolsep}{1pt}
\begin{tabularx}{\linewidth}{@{}l<{\hspace{4pt}}*{3}{>{\centering\arraybackslash}X}c@{}}
\toprule
Variants & \mbox{Stable~(\%) $\uparrow$} & \mbox{Drift~(mm) $\downarrow$} & \mbox{Lin. vel.~(m/s) $\downarrow$} & \mbox{Ang. vel.~(rad/s) $\downarrow$} \\
\midrule
SAM3D & \third 89.60 & \third 21.57 & \third 0.1997 & \third 4.8212 \\
\quad + RL & \second 89.92 & 23.03 & \second 0.1729 & \second 4.2923 \\
\quad + RL + IOA & 28.00 & 72.67 & 0.6786 & 10.3605 \\
\quad + RL + IOA + Center-RoPE & 89.58 & \second 19.05 & 0.2795 & 4.9019 \\
\quad \textbf{+ RL + IOA + Volume-RoPE} & \first 91.01 & \first 19.04 & \first 0.1637 & \first 3.8592 \\
\bottomrule
\end{tabularx}
\end{table}

\begin{table}[!htbp]
\caption{Ablation Study on \textbf{Object-wise GDPO} in Simulation.}
\vspace{-2mm}
\label{tab:stability_gdpo}
\centering
\small
\setlength{\tabcolsep}{1pt}
\begin{tabular*}{0.85\linewidth}{@{\extracolsep{\fill}}lcccc@{}}
\toprule
& \mbox{Stable~(\%) $\uparrow$} & \mbox{Drift~(mm) $\downarrow$} & \mbox{Lin. vel.~(m/s) $\downarrow$} & \mbox{Ang. vel.~(rad/s) $\downarrow$} \\
\midrule
Scene-Level GDPO & 85.33 & 28.34 & 0.2060 & 4.4171 \\
\textbf{Object-Wise GDPO} & \textbf{91.01} & \textbf{19.04} & \textbf{0.1637} & \textbf{3.8592} \\
\bottomrule
\end{tabular*}
\end{table}

\begin{table}[!htbp]
\caption{Ablation Study on \textbf{Physical-Plausibility Rewards} in Simulation.}
\vspace{-2mm}
\label{tab:stability_objective}
\centering
\footnotesize
\setlength{\tabcolsep}{1pt}
\begin{tabularx}{\linewidth}{@{}l<{\hspace{4pt}}*{4}{>{\centering\arraybackslash}X}@{}}
\toprule
Variants & \mbox{Stable~(\%) $\uparrow$} & \mbox{Drift~(mm) $\downarrow$} & \mbox{Lin. vel.~(m/s) $\downarrow$} & \mbox{Ang. vel.~(rad/s) $\downarrow$} \\
\midrule
SFT only & 41.38 & 46.44 & 0.3651 & 7.6000 \\
GT reward only & \second 82.57 & \second 25.00 & \second 0.3284 & \second 5.3486 \\
\textbf{Full objective} & \first 91.01 & \first 19.04 & \first 0.1637 & \first 3.8592 \\
\bottomrule
\end{tabularx}
\end{table}

In the simulation-based ablations, Volume-RoPE, object-wise GDPO, and reinforcement learning with physical-plausibility rewards each achieve the best results on all four stability metrics in their respective comparisons (\cref{tab:stability_ioa,tab:stability_gdpo,tab:stability_objective}).

\subsection{Attention Map Analysis}

\label{sec:attention_map}

\begin{figure*}
    \centering
    \includegraphics[width=\textwidth]{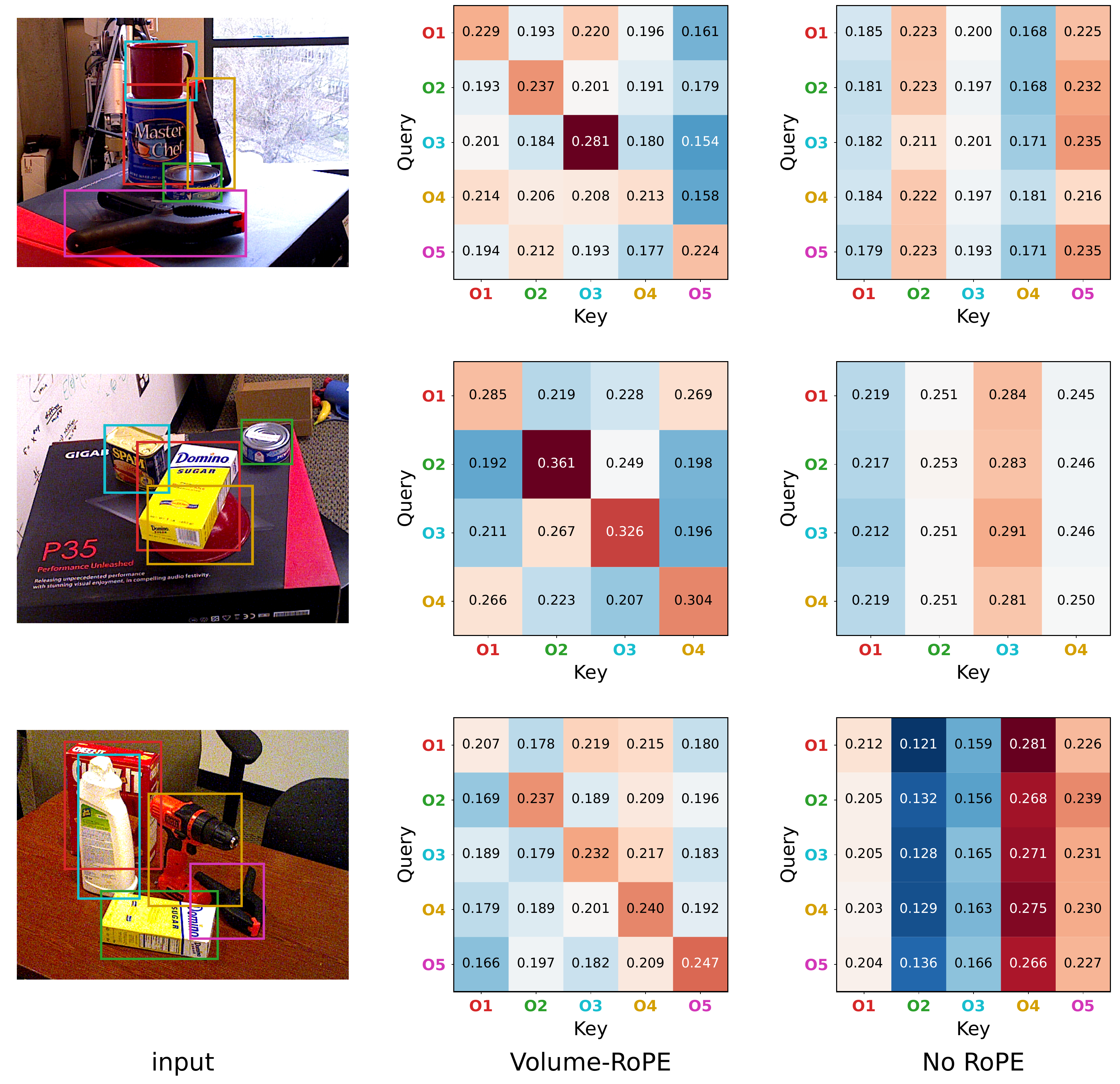}
    \captionof{figure}{\textbf{Average Inter-Object Attention Maps.}
    Each attention value is averaged across heads and layers, and the attention weights are summed over all tokens associated with each object.
    These attention maps are sampled at the first time step.
    Each object is matched to its corresponding query and key using its bounding box color.
    Red regions indicate higher attention weights, whereas blue regions indicate lower weights.
    }
    \vspace{-5mm}
    \label{fig:attention_map}
\end{figure*}

\Cref{fig:attention_map} illustrates how the objects attend to one another.
In the Volume-RoPE attention maps, each object generally focuses on nearby objects, although this pattern does not hold in every case.
For example, in the first input, object 5 focuses on object 2, its nearest neighbor.
Object 4 attends to objects 1--4 while assigning less attention weight to object 5, which is relatively far away.
In the second input, objects 1 and 4 attend to each other as object 1 is supported by object 4 in the input.
Finally, in the third input, object 1 assigns high attention weights to objects 3 and 4, which are also close to object 1.
In contrast, in the attention maps of the No RoPE variant, all queries tend to attend to same keys regardless of their spatial arrangement.
Together with the results in \cref{tab:ablation}, this observation suggests that the model has difficulty attending to the appropriate object without an explicit spatial cue such as Volume-RoPE.

\subsection{Coordinate Normalization in SAM3D}
\label{sec:use_scene_scale}

Because our experiments use a SAM3D normalization setting that differs from the default in the official repository, we clarify this distinction here.
In the official SAM3D implementation, the conditioning pointmap is normalized using the scale of the entire scene, such that the predicted scale and translation are expressed in this scene-level coordinate system.
In our setup, however, normalizing the input point map independently for each object yields substantially better performance, as shown in \cref{tab:results}, where \textit{repo ver.} denotes the original scene-normalized implementation, while the standard SAM3D row uses object-wise normalization.
We also find that SAM3D trains more robustly with reinforcement learning when using the coordinate system induced by object-wise normalization.
This normalization scheme is not our contribution; it is already supported by the original SAM3D code through the setting $use\_scene\_scale=false$.

\end{document}